%% file: main.tex
\pdfoutput=1
\documentclass{article}

\usepackage[margin=1in]{geometry}
\usepackage{times}
\usepackage{amsmath,amssymb,amsfonts}
\usepackage{booktabs}
\usepackage{graphicx}
\usepackage{hyperref}
\usepackage[capitalize,noabbrev]{cleveref}
\usepackage{natbib}
\usepackage{enumitem}
\usepackage{microtype}

\hypersetup{
  pdftitle={PolyGraph: Plan-Buffered CUDA Graphs for Adaptive Speculative Decoding},
  pdfauthor={Junchao Yang},
  hidelinks,
}

\input{math_commands.tex}

\title{PolyGraph: Plan-Buffered CUDA Graphs\\for Adaptive Speculative Decoding}

\author{Junchao Yang\thanks{Email: \texttt{yangjc27@gmail.com}}\\
  Hunan University}

\date{}

\begin{document}

\maketitle

\begin{abstract}
\input{sections/0_abstract.tex}
\end{abstract}

\input{sections/1_introduction.tex}
\input{sections/2_background.tex}
\input{sections/3_design.tex}
\input{sections/4_evaluation.tex}
\input{sections/6_conclusion.tex}

\small
\bibliographystyle{plainnat}
\bibliography{references}

\end{document}

%% file: math_commands.tex
\newcommand{\dmax}{D_{\max}}

\newcommand{\polygraph}{\textsc{PolyGraph}}

%% file: sections/0_abstract.tex
Adaptive speculative decoding changes the active batch size $B$ and draft length $d$ from one decode step to the next, whereas a CUDA Graph replays a fixed launch configuration.
The resulting verification width $M=Bd$ forces serving systems to pre-capture many shapes, capture new ones on the serving path, or pad every step to the largest draft.
We present \polygraph{}, a plan-buffer mechanism for custom Triton kernels that avoids this choice.
\polygraph{} captures one maximum-grid graph and writes the active shape to GPU memory before each replay.
Kernels skip tiles beyond that shape, while graph topology, allocations, layouts, pointers, and launch grids remain fixed; the update transfers only 28 bytes.
On one NVIDIA H800, a trace-driven prototype with Qwen2.5-3B dimensions replaces 64 shape-specific graphs with one, matches the warmed baseline in observed step latency (17.7 vs.\ 17.8\,ms P50), and reduces measured pre-serving graph-build time from 3{,}915 to 93\,ms.
Fixed-draft controls show that the benefit comes from avoiding padded payload: the P99 difference from MaxPad is negligible at zero padding waste but reaches an 18.7\% reduction at 100\% waste.
Our evidence establishes a bounded work-shape mechanism for custom kernels, not a production serving system, and is limited to synthetic traces on one GPU.

%% file: sections/1_introduction.tex
\section{Introduction}
\label{sec:intro}

Speculative decoding replaces several target-model steps with one parallel verification pass: a smaller model drafts a token block, and the target accepts a prefix~\citep{leviathan2023fast,chen2023accelerating}.
The best block length varies across inputs and decoding steps.
When predictions are easy, a longer draft amortizes verification over more accepted tokens; after an early rejection, however, its extra positions are wasted.
Adaptive methods therefore choose a new draft length at each step~\citep{liu2024pearl,agrawal2024adaedl,wang2024opttree}.
Meanwhile, continuous batching changes the active request set~\citep{yu2022orca}.
Together, these policies give the target model a token dimension $M=Bd$ that varies with both batch size $B$ and batch-wide draft length $d$.

CUDA Graphs work best under the opposite assumption.
They remove repeated CPU launch overhead by recording a GPU workload once, but capture also fixes kernel grids and arguments.
A serving system can accommodate a changing $M$ in three conventional ways.
\emph{AllShapes} captures every supported shape before serving and pays in startup time and graph count.
\emph{OnDemand} captures each observed shape only when it first appears, causing a first-use latency spike.
\emph{MaxPad} keeps a smaller graph set by always running the largest draft, even when most of those token positions will be discarded.

\polygraph{} keeps the captured workload fixed without treating the entire maximum shape as useful work.
It stores the active dimensions in a small GPU-resident \emph{plan buffer} whose address, unlike its contents, is fixed during capture.
Before each replay, the runtime updates seven plan fields with a 28-byte asynchronous copy.
Every plan-aware Triton kernel then reads its active extent and returns from tiles that fall outside it.
The same graph can serve every $M$ within its captured bound, while inactive positions pay for a guard rather than MaxPad's full kernel payload.

We evaluate the mechanism in a trace-driven prototype on one NVIDIA H800.
For a 36-layer workload with Qwen2.5-3B dimensions, \polygraph{} uses one graph where AllShapes uses 64 and MaxPad uses 16.
It cuts measured pre-serving graph-build time from 3{,}915 to 93\,ms and matches the warmed reference in observed latency (17.7 vs.\ 17.8\,ms P50).
OnDemand instead exposes first-use captures in its tail, while MaxPad spends 31\% of token work on padding.
Fixed-draft controls isolate that padding cost: \polygraph{} differs from MaxPad by only 0.3\% when $d=d_{\max}$, but lowers P99 by 18.7\% when padding doubles the executed token count.

\pagebreak[3]
We make three contributions:
\begin{enumerate}[leftmargin=*,itemsep=2pt]
    \item We formulate adaptive draft length as a dynamic CUDA-Graph work dimension and quantify the costs of shape enumeration, online capture, and maximum padding.
    \item We introduce a graph-wide plan-buffer contract that lets custom kernels vary their bounded active extent while preserving a fixed captured workload.
    \item We compare the mechanism with all three alternatives and use fixed-draft controls to connect its performance directly to avoided padding.
\end{enumerate}

Our claims stop at this mechanism boundary.
The prototype replays custom Triton kernels over synthetic serving traces; it does not integrate \polygraph{} into vLLM or any other production engine.

%% file: sections/2_background.tex
\section{Background and Related Work}
\label{sec:background}

\subsection{Adaptive Draft Length}

In speculative decoding, a draft model proposes $d$ tokens and the target model verifies the block in parallel~\citep{leviathan2023fast,chen2023accelerating}.
The target accepts only a prefix, so increasing $d$ trades more opportunities for acceptance against more work after the first rejection.
PEARL adjusts the drafting performed alongside verification~\citep{liu2024pearl}; AdaEDL stops drafting from an entropy signal~\citep{agrawal2024adaedl}; and OPT-Tree selects a candidate tree for each step~\citep{wang2024opttree}.
Although their policies differ, all three turn verification width into a runtime decision.

Continuous batching changes that width again by admitting and retiring requests after each iteration~\citep{yu2022orca}.
When a batch uses a rectangular verification tensor, it executes the largest draft requested by any active sequence, $d=\max_i d_i$.
The resulting token dimension $M=Bd$ is known only after the scheduler has assembled the step.

\subsection{CUDA Graph Strategies}
\label{sec:bg-strategies}

CUDA Graphs record a sequence of GPU operations and replay it without relaunching every kernel from the CPU~\citep{nvidia2024cudagraph}.
The recorded executable includes kernel grids, pointers, and scalar arguments, so a new shape normally requires a different graph, an executable update, or eager execution.
Serving engines such as vLLM~\citep{kwon2023vllm} and SGLang~\citep{zheng2024sglang} already manage a dynamic batch axis by capturing selected sizes and padding each batch to a supported shape.
Adaptive drafting introduces another axis that can change at every decode step.

\Cref{tab:strategy-comparison} compares four ways to manage this shape space.
AllShapes pays up front to capture every supported bucketed shape.
OnDemand instead keys captures by the exact observed $M$ and context length, so the first request for each new pair waits for capture.
MaxPad removes $d$ from the graph key by executing $B d_{\max}$ positions at every step, which adds $d_{\max}/d-1$ relative payload work.
\polygraph{} instead captures $K$ maximum-grid buckets and guards the inactive tiles; we evaluate the simplest case, $K=1$.

\begin{table}[h]
\centering
\caption{Strategies for the dynamic token dimension $M=Bd$. $K$ denotes the number of maximum-grid buckets.}
\label{tab:strategy-comparison}
\vspace{3pt}
\setlength{\tabcolsep}{5pt}
\begin{tabular}{@{}lccc@{}}
\toprule
Strategy & Graphs & Capture time & Inactive-token work \\
\midrule
AllShapes & $O(B_{\max}d_{\max})$ & Before serving & None \\
OnDemand  & $O(\text{shapes seen})$ & During serving & None \\
MaxPad    & $O(B_{\max})$ & Before serving & $d_{\max}/d-1$ \\
\textbf{\polygraph{}} & $\boldsymbol{O(K)}$ & Before serving & Tile guards \\
\bottomrule
\end{tabular}
\end{table}

\subsection{Device-Side Indirection}
\label{sec:related}

Device-side metadata already decouples operator execution from values computed on the host.
FlashInfer separates attention planning from execution~\citep{ye2025flashinfer}; CUB precomputes metadata for variable-size primitives~\citep{merrill2015cub}; and grouped-GEMM kernels read token counts on the device to skip padded expert blocks~\citep{vllm2024fusedmoe,deepseek2025deepgemm}.
PyGraph removes repeated host-to-device parameter copies during graph replay~\citep{ghosh2025pygraph}.
\polygraph{} builds on these ideas rather than claiming device-side counts as a new primitive.
Its contribution is a shared shape plan that governs every custom kernel in a captured decode graph, together with an evaluation of that contract under adaptive speculative-decoding shapes.

%% file: sections/3_design.tex
\section{\polygraph{} Design}
\label{sec:design}

\polygraph{} moves shape-dependent values out of captured kernel arguments and into persistent device memory.
The graph always launches the grid associated with a bucket maximum; a per-step plan tells each kernel how much of that grid should perform payload work.

\subsection{Plan Buffer}
\label{sec:plan-buffer}

The plan buffer occupies eight \texttt{int32} slots (32 bytes) in a persistent GPU allocation.
The 28-byte update writes seven active slots, where $H$ and $I$ are the hidden and intermediate widths, $N_h$ is the number of attention heads, and $s$ is the KV-cache slot base.
Slot 7 is reserved.
The evaluated graph-selection axis is $M$; context length also varies within its captured bound, while the cache entry remains scheduler metadata.
Capture fixes the buffer's address, not its values.

\begin{table}[h]
    \centering
    \small
    \renewcommand{\arraystretch}{0.92}
    \setlength{\tabcolsep}{9pt}
    \caption{The seven active fields copied before each replay.}
    \label{tab:plan-layout}
    \begin{tabular}{cll}
        \toprule
        Slot & Value & Meaning \\
        \midrule
        \texttt{p[0]} & $M=Bd$ & token rows \\
        \texttt{p[1]} & $MH$ & hidden elements \\
        \texttt{p[2]} & $MI$ & MLP elements \\
        \texttt{p[3]} & $C$ & context length \\
        \texttt{p[4]} & $MN_h$ & token--head pairs \\
        \texttt{p[5]} & $d$ & draft length \\
        \texttt{p[6]} & $s$ & KV-cache slot base \\
        \bottomrule
    \end{tabular}
\end{table}

Before replay, the runtime writes the fields to pinned host memory and enqueues one 28-byte \texttt{cudaMemcpyAsync} on the graph's launch stream.
Stream ordering exposes the new plan before any replayed kernel reads it.
On our H800, the update takes approximately 3.6\,$\mu$s, compared with roughly 18\,ms for a 36-layer step.
When the host queues ahead, an event-guarded staging ring prevents reuse while a copy remains in flight.

\begin{figure*}[t]
    \centering
    \includegraphics[width=0.95\textwidth]{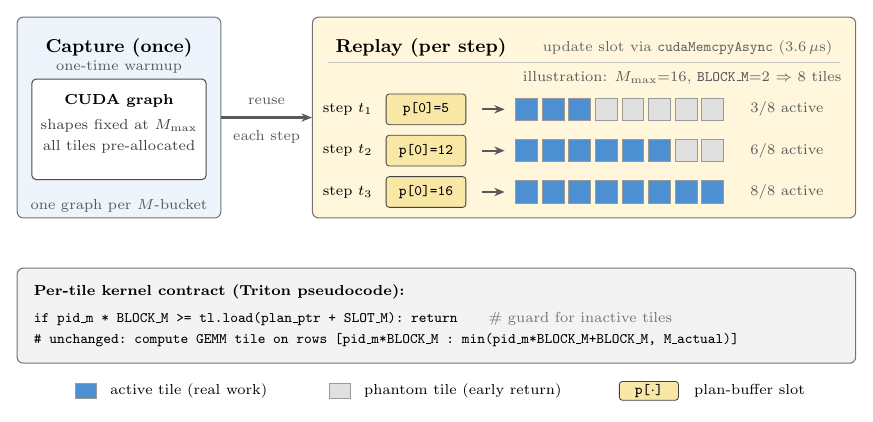}
    \caption{Plan-buffer replay. Capture fixes a bucket's maximum grid and the plan-buffer address. Before each replay, a 28-byte copy updates the seven active fields. Every Triton program instance reads its extent from the plan: active tiles run the original payload, while tiles beyond the extent return after the guard.}
    \label{fig:plan-buffer}
\end{figure*}

\subsection{Kernel Contract and Dispatch}
\label{sec:dispatch}

A plan-aware kernel replaces each captured scalar extent with a load from the plan buffer.
For a kernel tiled along the token axis, the essential guard is
\[
    \texttt{if tile\_start >= load(plan[M]): return}.
\]
Tiles within the active extent execute the original computation.
Tiles beyond it pay only for the metadata load and comparison; they do not run the matrix multiplication or other payload.
This contract applies to kernels whose work can be bounded by a device-side guard, but not automatically to opaque library calls with an inaccessible launch shape.

For each decode step, the runtime computes $M=Bd$ and selects the smallest captured bucket that covers it.
It then updates the plan and replays the corresponding graph.
Our evaluation uses one bucket captured at $M_{\max}=B_{\max}d_{\max}$, so every supported step shares the same launch grid.
Additional buckets could trade a larger graph set for fewer inactive guards, but we leave that tradeoff outside the present study.

\subsection{Guarding versus Padding}

MaxPad and \polygraph{} both replay a graph captured at a maximum shape; they differ in what happens inside that graph.
MaxPad marks all $B d_{\max}$ positions as valid, causing projections, normalization, and feed-forward layers to process tokens that the runtime will discard.
\polygraph{} marks only the actual $Bd$ positions as active, so the remaining tiles stop at the guard.
The performance difference should therefore track the amount of padded payload.
When $d=d_{\max}$, both methods execute the same useful token count; as $d$ falls, MaxPad performs progressively more discarded work.
\Cref{sec:eval-decomposition} tests this prediction with fixed-draft controls.

%% file: sections/4_evaluation.tex
\section{Evaluation}
\label{sec:eval}

Our evaluation addresses three questions.
Can one \polygraph{} graph retain the latency of a warmed shape-specific cache?
Does it avoid the capture misses of an online cache?
And does its advantage over maximum padding grow with the work that padding introduces?

\subsection{Setup}
\label{sec:eval-setup}

We run all experiments on one NVIDIA H800 PCIe GPU.
The prototype executes a 36-layer decode workload with Qwen2.5-3B dimensions and custom Triton kernels, but no trained weights.
The serving space contains batch sizes $B\in\{1,\ldots,16\}$ and draft lengths $d\in\{1,\ldots,8\}$.
These 128 $(B,d)$ pairs produce 64 distinct values of $M=Bd$.
AllShapes captures one graph for each of those 64 values.
MaxPad keeps the exact batch size but fixes $d=8$, yielding 16 graphs.
\polygraph{} guards the combined token extent and uses one maximum-grid graph.

No public serving trace records the per-step draft lengths required by this study, so we generate request traces.
The primary trace combines Poisson arrivals, ShareGPT-calibrated sequence lengths, and a PEARL-style schedule that shortens the draft as generation proceeds.
For each of three seeds (42, 1234, and 2025), the benchmark serves 400 requests and sets the step width to the largest active draft, $d=\max_i d_i$.

We measure step latency, graph count, pre-serving graph-build time, and payload-padding waste.
Graph-build time measures wall time around graph construction, not CUDA capture in isolation, and can include compilation first triggered during capture.
AllShapes completes its entire graph set before measurement and serves as a warmed performance reference.
OnDemand begins with an empty cache and captures on a miss, whereas MaxPad always executes $d_{\max}$.
Every strategy replays the same request trace for a given seed.

\subsection{Serving Comparison}
\label{sec:eval-serving}

\Cref{tab:e2-serving} averages the per-seed latency quantiles.
\polygraph{} closely tracks the warmed AllShapes reference: their P50 latencies are 17.7 and 17.8\,ms, and both P99 values round to 18.4\,ms.
It reaches that latency with one graph and 93\,ms of pre-serving graph building, compared with 64 graphs and 3{,}915\,ms for AllShapes.

The other baselines expose the costs that \polygraph{} is designed to avoid.
OnDemand removes pre-serving construction but raises P99 to 62.0\,ms because graph misses occur on the serving path.
MaxPad avoids those misses, yet padding accounts for 31\% of its token work and raises P50 to 18.4\,ms.
For this custom-kernel workload, plan-buffer replay thus reaches the observed latency of a fully warmed shape-specific cache without maintaining that cache's graph set.

\input{figures/TABLE_e2_serving.tex}

\subsection{Padding-Waste Decomposition}
\label{sec:eval-decomposition}

To isolate padding, we repeat the benchmark with fixed drafts $d\in\{1,4,8\}$ over the same three seeds.
With $d_{\max}=8$, these settings make MaxPad execute 700\%, 100\%, and 0\% more token positions than required.
The P99 results in \Cref{tab:e7-decomposition} follow that ordering.
At $d=8$, both methods execute the same token width, and \polygraph{}'s P99 is only 0.3\% lower.
At $d=4$, MaxPad doubles the required token work, and \polygraph{} reduces P99 from 18.4 to 15.0\,ms ($-18.7\%$).
At $d=1$, the reduction grows to 33.4\%.
Across seeds, the sample standard deviation of these relative gaps never exceeds 0.12 percentage points.
The gain is therefore conditional: plan-buffer replay helps when it removes padding and offers little benefit when the workload already fills the captured shape.

\input{figures/TABLE_e7_decomposition.tex}

\subsection{Scope and Limitations}
\label{sec:limitations}

\paragraph{Prototype rather than serving-engine integration.}
The benchmark replays custom Triton kernels over synthetic traces.
It excludes a production scheduler, networking and tokenization, paged KV-cache management, and the adaptive draft model itself.
AllShapes therefore establishes mechanism-level latency parity; the experiment does not establish an end-to-end speedup in vLLM.

\paragraph{Plan-aware kernels.}
\polygraph{} requires every shape-dependent kernel to read the plan and guard inactive tiles.
Kernels controlled by the serving system can adopt this contract directly, but opaque cuBLAS or third-party calls cannot.
A production engine would need plan-aware library interfaces or graph boundaries around incompatible operators.

\paragraph{Evaluation coverage.}
All measurements come from one H800 and one model-shaped configuration, with synthetic arrival and draft schedules.
Broader deployment claims require trained-model execution, public production traces, additional GPUs, and integration into a complete serving engine.

%% file: figures/TABLE_e2_serving.tex
\begin{table*}[!t]
\centering
\caption{Trace-driven comparison on an H800 (Qwen2.5-3B dimensions, 36 layers, 400 requests). We average the quantile from each of seeds 42, 1234, and 2025; all times are in ms. Pre-build measures benchmark wall time around graph construction. $\dagger$AllShapes is fully built before serving, whereas OnDemand captures on a miss.}
\label{tab:e2-serving}
\setlength{\tabcolsep}{5pt}
\begin{tabular}{l rrr rrr}
\toprule
& \multicolumn{3}{c}{Step latency (ms)} & \multicolumn{3}{c}{Resources} \\
\cmidrule(lr){2-4} \cmidrule(lr){5-7}
System & P50 & P90 & P99 & Graphs & Pre-build & Payload waste \\
\midrule
AllShapes$\dagger$    & 17.8 & 18.4 & 18.4 & 64 & 3915 & 0\% \\
OnDemand              & 17.8 & 18.4 & 62.0 & 39 & 0 & 0\% \\
MaxPad                & 18.4 & 18.4 & 18.4 & 16 & 911 & 31\% \\
\textbf{\polygraph{}} & \textbf{17.7} & \textbf{18.4} & \textbf{18.4} & \textbf{1} & 93 & \textbf{0\%} \\
\bottomrule
\end{tabular}
\end{table*}

%% file: figures/TABLE_e7_decomposition.tex
\begin{table}[tb]
\centering
\caption{Fixed-draft controls averaged over seeds 42, 1234, and 2025; $\pm$ denotes the sample standard deviation across seeds. \polygraph{}'s P99 reduction relative to MaxPad grows with the payload that padding adds.}
\label{tab:e7-decomposition}
\setlength{\tabcolsep}{6pt}
\begin{tabular}{l r rr r}
\toprule
Scenario & MaxPad waste & MaxPad P99 (ms) & PG P99 (ms) & P99 Gap \\
\midrule
fixed=8 ($\dmax{}$) & 0\%   & 18.4 & 18.3 & $-0.3\pm0.03\%$ \\
fixed=4             & 100\% & 18.4 & \textbf{15.0} & $\mathbf{-18.7\pm0.06\%}$ \\
fixed=1             & 700\% & 18.4 & \textbf{12.2} & $\mathbf{-33.4\pm0.12\%}$ \\
\bottomrule
\end{tabular}
\end{table}

%% file: sections/6_conclusion.tex
\section{Conclusion}
\label{sec:conclusion}

Adaptive draft length changes the verification shape at the same cadence that a serving system would like to replay a CUDA Graph.
\polygraph{} reconciles the two by keeping the captured workload fixed and placing the active token extent in a GPU-resident plan.
In our trace-driven H800 prototype, this contract replaces a warmed set of 64 shape-specific graphs with one graph, preserves similar observed step latency, and reduces measured pre-serving graph-build time from 3{,}915 to 93\,ms.
Fixed-draft controls show where the improvement comes from: guarding inactive tiles matters in proportion to the payload work that maximum padding would otherwise execute.

These results support a bounded mechanism for plan-aware custom kernels, not a complete serving architecture.
The next step is to integrate the same contract into an adaptive speculative-decoding engine and evaluate it with trained models and production traces.